\documentclass{WileyMSP-template}
\usepackage{caption}
\usepackage{xcolor}
\usepackage{multirow}
\usepackage{graphicx}
\usepackage{float} 
\usepackage{subfigure}
\usepackage{amsmath}
\usepackage{mathtools}
\usepackage{amssymb}
\usepackage{unicode-math}
\usepackage{makecell}
\usepackage[colorlinks, linkcolor=black, menucolor=black, citecolor=black, urlcolor=blue]{hyperref}
\usepackage{url}
\hypersetup{
    colorlinks=true,
    linkcolor=black,
    filecolor=black,      
    urlcolor=blue,
    citecolor=black,
}

\newcounter{refcounter} 

\begin{document}

\pagestyle{fancy}
\rhead{\includegraphics[width=2.5cm]{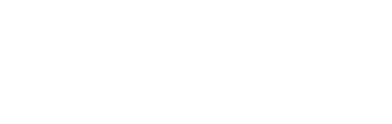}}

\title{Learning Highly Dynamic Skills Transition for Quadruped Jumping Through Constrained Space}

\maketitle


\author{Zeren Luo$^{*1}$}
\author{Jiahui Zhang$^{*1}$}
\author{Yimin Han$^{1}$}
\author{Ji Ma$^{1}$}
\author{Minghao Lu$^{1}$}
\author{Ioannis Havoutis$^{2}$}
\author{Peng Lu$^{\dagger1}$}



\begin{affiliations}
* Equal contribution. $^{\dagger}$Corresponding author: \url{lupeng@hku.hk} \\

$^{1}$Adaptive Robotic Controls Lab (ArcLab), Department of Mechanical Engineering, University of Hong Kong, Hong Kong SAR, China. \\

$^{2}$Dynamic Robots Systems Group (DRS), Oxford Robotics Institute, Department of Engineering Science, University of Oxford, U.K, \url{ioannis@robots.ox.ac.uk} \\

This work was supported by the General Research Fund under Grant 17204222, and in part by the Seed Fund for Collaborative Research and General Funding Scheme-HKU-TCL Joint Research Center for Artificial Intelligence.

\end{affiliations}


\keywords{Legged Robots, Highly Agile Motions, Deep Reinforcement Learning}

\begin{abstract}

Although legged animals are capable of performing explosive motions while traversing confined spaces, replicating this behavior in quadrupedal robots has been a longstanding challenge. Here, we propose a hierarchical reinforcement learning pipeline that empowers the robots to perform aggressive locomotion through constrained obstacles--a narrow gate. The imitation learning technique is used to train the low-level policy, which mimics the behaviors of real animals and forms a set of diverse skills. The high-level controller, having an awareness of the capability of low-level skills and acquiring the gate information via vision-based detection, determines the suitable maneuvers with collision-free trajectories to traverse it dynamically.
Notably, we also verify that this framework can be extended to other highly dynamic tasks.
This is one of the first works that perform autonomous and agile aerial gate traversal tasks on ground-walking robots, extending the lifelike agility of legged robots to match that of their biological counterparts. 

\end{abstract}


\section{Introduction}
\begin{figure}[t]
   \centering
   \vspace{+0.7em}
    \includegraphics[width=0.56\textwidth, height=0.28\textwidth, trim=1 1 1 1,clip]{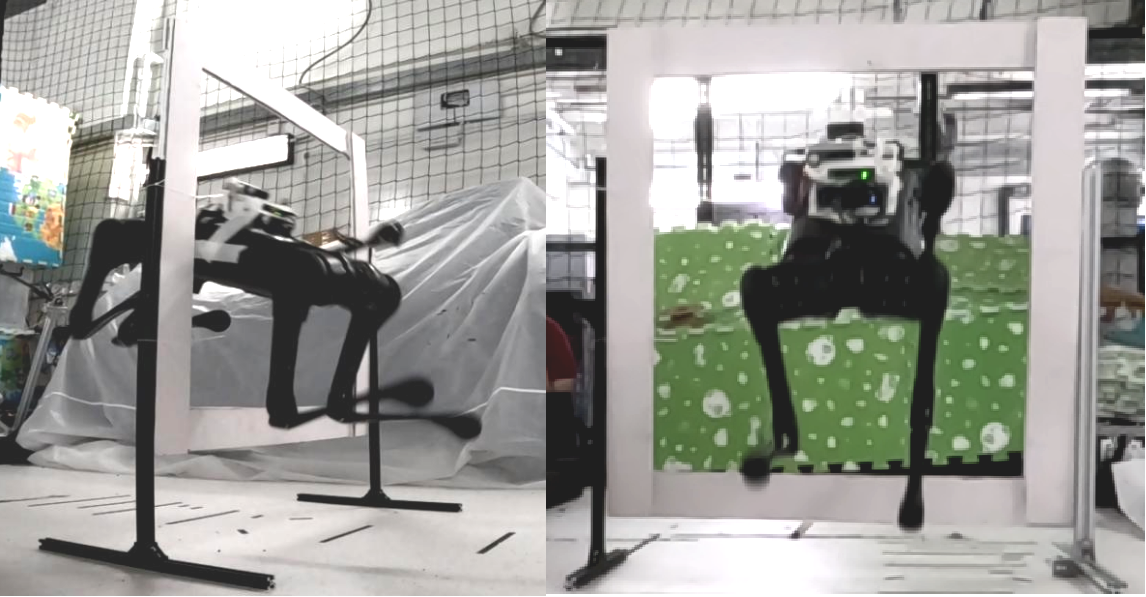}
   \caption{ \textcolor{black}{Legged robots require highly aggressive locomotion skills and perception when they jump through a narrow space that is comparable with their body size.}
   }
   \label{pics:cover}
\end{figure}

In natural environments, animals that excel in dynamic movements, such as kangaroos, frogs, and dogs exhibit remarkable agility such as high-speed jumping and sprinting. 
It has been challenging to reproduce these behaviors on quadrupedal robots. At high speeds, the robot's dynamics are affected by multiple physical factors, such as motor limits, contact force modulation, and maneuverability in flight phases. 
Additionally, animals can perform other downstream tasks under high-dynamic scenarios, such as accurately jumping through confined spaces, as shown in Fig. \ref{pics:cover}. Legged robots are still far from fully exploiting hardware's capabilities to achieve similar behaviors, as it is highly demanding not only the robot's athleticism but also its decision-making abilities, without which it would be impossible to determine the optimal take-off time and navigate through narrow spaces.

\subsection{Locomotion with Highly Dynamic Behavior} 
Traditional methods rely on trajectory optimization (TO) [\ref{nguyen2019optimized}, \ref{Park_2015}, \ref{ding2024robust}] to generate aerial motion for legged robots. \textcolor{black}{For standing jumps, full dynamics results in time-consuming optimizations}, restricting TO to offline solutions [\ref{nguyen2019optimized}]. More demanding running jumps require real-time optimal trajectories solved with simplified models [\ref{Park_2015}, \ref{li2024cafempc}] where their performance can be affected by violations of physical constraints.
Several recent studies focus on hardware-level modifications to enhance jumping performance, which involves incorporating flywheels for better controllability in the aerial phase [\ref{roscia2023orientation}] and springs on actuated joints to improve robustness to uncertainty [\ref{ding2024robust}].

Learning-based methods have been considered as an alternative line for generating highly dynamic behaviors [\ref{cheng2023extreme}]. Early studies utilize reinforcement learning (RL) to generate directly the optimal jumping trajectory [\ref{margolis2022learning}] or the compensatory adjustments [\ref{bellegarda2020robust}]. Nevertheless, they are constrained by the predefined gait, allowing only specific bouncing patterns [\ref{yang2023cajun}]. Triggering the jumping behavior from scratch requires demanding reward and curriculum design [\ref{atanassov2024curriculum}]. 
Animal motions, however, possess inherent naturalistic properties that help enhance the robot's learned style.
With the recent advances in computer graphics and robotics, it is now possible to replicate agile movements from legged animals [\ref{grandia2023doc}, \ref{zhang2018mode}], and generate life-like jumping on real robots via imitation learning [\ref{zhang2024learning}, \ref{smith2023learning}]. In this work, we also mimic highly dynamic animal maneuvers and versatile skills to compose a unified locomotion policy.

\subsection{Hierarchical RL for Legged Robots} 
Conventional approaches of hierarchical RL typically decouple the low-level from the high-level controller, which becomes agnostic to the proprioceptive states related to the locomotion [\ref{kareer2023vinl}].
Follow-up research reuses the locomotion policy and proprioception [\ref{bohez2022imitate}] to train task-specific high-level policy, which then controls a collection of motion primitives.
In these approaches, a high-level controller learns to switch between discrete subpolicies [\ref{yang2020multi}, \ref{hoeller2024anymal}]. However, this discontinuity potentially leads to overlooking some intermediate states.
Low-level maneuvers can also be integrated implicitly into a unified policy either by distillation [\ref{caluwaerts2023barkour}] or are embedded into a latent space [\ref{peng2022ase}], while the distilled policy's performance often struggles to outperform the undistilled version and some are only validated on animation characters.

In this work, the unified locomotion policy synthesizes the skills as continuous states, and then a high-level policy leverages these skills to guide the quadrupedal robot to perform a novel downstream task.

\subsection{Narrow Space Traversal} 

Aggressive flight through the narrow gap for UAVs has been extensively explored for decades [\ref{mellinger2012trajectory}, \ref{kaufmann2023champion}]. Though drones have relatively simpler dynamics than legged robots, acquiring real-time decision-making capability for high-speed agile maneuvers remains a challenging task, making it the ultimate test of aerial vehicle performance. However, until now, few studies have empowered the legged robots to dynamically traverse such narrow spaces with an aerial phase. 
Previous parkour jumping simply demands landing on a suitable spot [\ref{cheng2023extreme}, \ref{zhuang2023robot}],
but it becomes challenging when aerial constraints significantly limit plausible movements.

A few related examples of legged robots traversing restricted spaces include walking under low obstacles using special crouching gaits [\ref{li2023autonomous}, \ref{kim2024not}] or squeezing through thin slits [\ref{zhuang2023robot}]. The constantly exerted contact forces ensure that the robot is controllable and operates in a relatively uniform pattern. However, with regard to traversing narrow gaps in the air, the robot needs to transit among a wider range of phases [\ref{gilroy2021autonomous}] including periods of complete aerial movement where control is extremely limited due to the absence of contact forces (Fig. \ref{pics:cover}).

In summary, we propose a hierarchical RL framework for quadrupedal robots to achieve high-dynamic locomotion and constrained space traversal tasks (ConsJump). The controller trained by the framework enables the robot to jump through a gap of comparable size to the robot itself. The key contributions of this work are listed as follows:
\begin{itemize}
    \item An imitation-based locomotion policy first integrates versatile agile maneuvers. A high-level controller, informed by vision-based detection, is then trained to exploit these maneuvers to generate feasible whole-body trajectories.
    \item Using the real-robot platform in Fig. \ref{pics:hardware_setup}, we demonstrate that the resulting system achieves lifelike agility by fully leveraging the motor limits. Reasoning about the surroundings, the robot can autonomously adapt to varying scenes.
    \item Further experiments are conducted to validate that such hierarchical architecture can be easily extended to novel tasks by reconfiguration of the high-level decision module. 
    
\end{itemize}

To the best of our knowledge, this work is one of the first examples demonstrating autonomous and highly dynamic narrow space traversal on legged robots and validating on real hardware.

\section{Methodology}

\subsection{Problem Statement}
\begin{figure*}[h]
   \centering
    \includegraphics[width=1.005\textwidth]{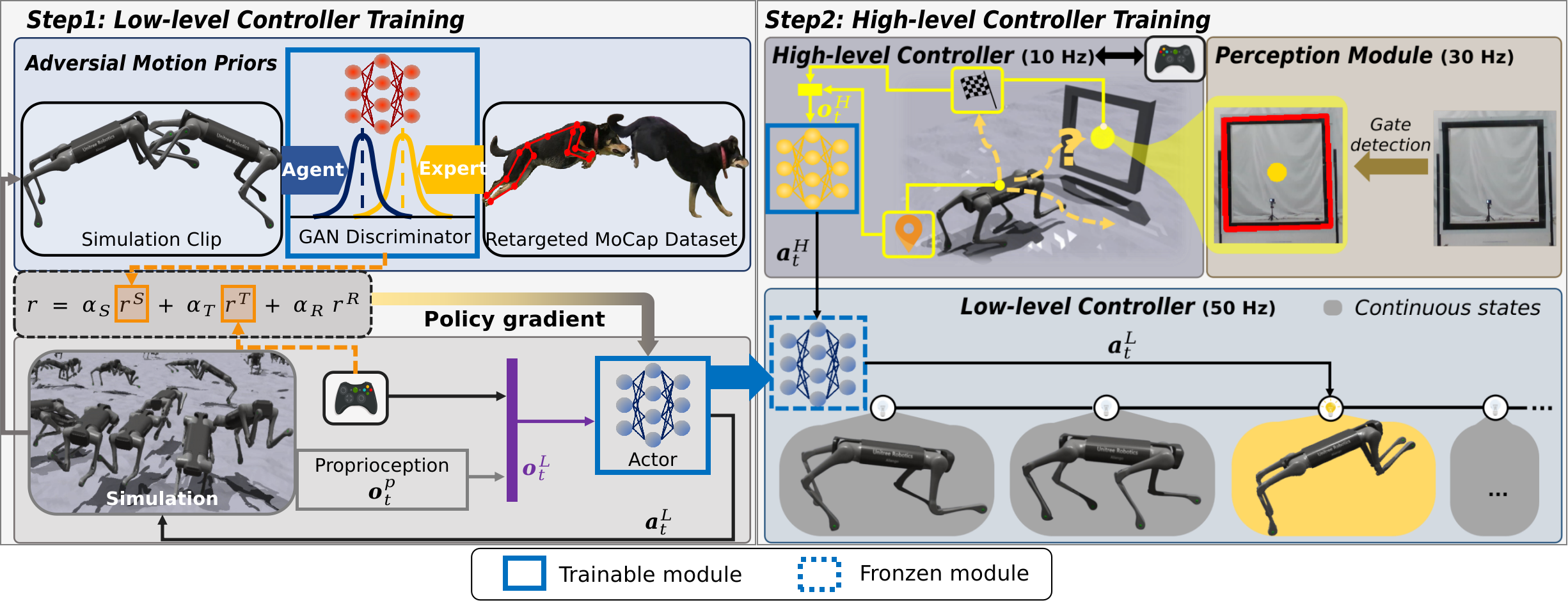}
   \caption{The schematic of the proposed training methodology. In step 1, a low-level locomotion policy is trained using adversarial RL, which guides the learning agent to track the command velocity as well as imitate the motion primitives. \textcolor{black}{In step 2, the locomotion policy is frozen, and a high-level policy is trained to generate the command velocity at a lower frequency.} The high-level module receives visual detection results to identify the gate information. When the command actions provided by the high-level policy change, the low-level policy implicitly generates the corresponding skills accordingly.  }
   \label{pics:framework}
\end{figure*}

Our pipeline follows a hierarchical structure, as shown in Fig. \ref{pics:framework}, and both the high-level and low-level modules are obtained through reinforcement learning (RL). The RL-based control problem can be formulated as a Markov Decision Process (MDP) defined by $\{ \mathcal{S}, \mathcal{A}, \mathcal{R}, \mathcal{P}, \gamma\}$, where $\mathcal{S}$ is the state space and $\mathcal{A}$ is the action space. At state $\boldsymbol{s}_t$ of time step $t$, a policy $\pi_{\theta}$ performs an action $\boldsymbol{a}_t$ to forward the environment to the next state $\boldsymbol{s}_{t+1}$ with transition probability $\mathcal{P}(\boldsymbol{s}_{t+1} | \boldsymbol{s}_{t}, \boldsymbol{a}_t)$ meanwhile receiving reward $\mathcal{R}(\boldsymbol{s}_t, \boldsymbol{a}_t)$ ($r_t$ for abbreviation). RL aims to find the optimal parameter $\theta$ that maximizes the discounted return: $J(\theta) = \mathbb{E}_{\pi_{\theta}}\left[\sum_{t=0}^{\infty}(\gamma^{t}r_{t})\right]$. Consistent with this problem formulation, we provide detailed descriptions of each RL component of our system in the following sections.

\subsection{Training of Low-Level Controller} \label{low_level}
In the first stage, a locomotion controller is trained to track omnidirectional velocity commands and model a versatile repertoire of skills by imitating the animal behaviors from a large unstructured dataset. \textcolor{black}{The resulting policy $\pi^{L}$ generates joint-level actions $\boldsymbol{a}^{L}_{t}$ to control the robots.}

\subsubsection{State space} \label{low_level_state} The input to the low-level policy $\boldsymbol{o}^{L}_{t}$ consists of proprioception $\boldsymbol{o}^{p}_{t}$ and commanded velocity $\boldsymbol{v}^{cmd}_{t}$, as shown in Fig. \ref{pics:framework}. $\boldsymbol{o}^{p}_{t} \in\mathbb{R}^{42}$ contains base angular velocity $\boldsymbol{\omega}_{t} \in\mathbb{R}^{3}$, projected gravity $\boldsymbol{g}_{t} \in\mathbb{R}^{3}$, joint angles $\boldsymbol{q}_{t} \in\mathbb{R}^{12}$, joint velocities $\boldsymbol{\dot{q}}_{t} \in\mathbb{R}^{12}$, and the action of the last step $\boldsymbol{a}^{L}_{t-1} \in\mathbb{R}^{12}$. As the base linear velocity $\boldsymbol{v}_{t} \in\mathbb{R}^{3}$ cannot be directly acquired on the real machine, we consider $\boldsymbol{v}_{t}$ as part of the privileged states that serve as input to the critic net.

\subsubsection{Action space}
The action $\boldsymbol{a}^{L}_t \in\mathbb{R}^{12}$ is the desired offset of the joint angle added on a time-invariant nominal pose $\boldsymbol{\mathring{q}}$, i.e. $\boldsymbol{a}^{L}_{t} = \boldsymbol{q}^{*}_{t} - \boldsymbol{\mathring{q}}$. The final desired angle $\boldsymbol{q}^{*}_{t}$ is tracked by the torque generated by the joint-level PD controller.
To mitigate the reality gap in real-world deployment, this motor actuation module is replaced by an actuator network trained with data collected from real machines, following the steps in our previous work [\ref{luo2024moral}].

\subsubsection{Reward function}
A generative imitation learning is adopted, which employs GAN-style training to optimize a “style” performance from a set of motion demonstrations $\mathcal{M}$ [\ref{peng2021amp}, \ref{escontrela2022adversarial}]. The procedure can be interpreted as the learning agent yielding the states and actions that the discriminator cannot reliably distinguish from the real data. \textcolor{black}{The reward engineering consists of different components involving the style $r^{S}_{t}$, the task $r^{T}_{t}$, and the regularization $r^{R}_{t}$ with respective ratios $\alpha$}:
\begin{equation}
r_t = \alpha_{T} \cdot r^{T}_{t} +  \alpha_{S} \cdot r^{S}_{t} + \alpha_{R} \cdot r^{R}_{t}
\label{eq:reward_total}
\end{equation}
The task-specific reward $r^{T}_{t}$ encourages the robot to track command velocity, as well as achieve higher foot heights during the aerial phase. The style reward $r^{S}_{t}$ encourages the similarity between the reference and the agent. 
\textcolor{black}{The training objective of the network-based discriminator $\mathcal{D}_{\psi}$ aims to assign higher scores for samples collected from reference $\mathcal{M}$ than that from policy $\pi^{L}$.
$r^{S}_{t}$ is calculated given the inference of $\mathcal{D}_{\psi}$.}
The regularization $r^{R}_{t}$ reward allows the robot to achieve better smoothness. More detailed terms can be found in Table \ref{reward_function}.

\subsubsection{Reference motion dataset} \label{dataset} \textcolor{black}{Inspired by biological systems where animals adapt their locomotion style based on acceleration and deceleration to jump through obstacles}, we select motion clips ranging from low to high-speed [\ref{zhang2018mode}] including pace, cantering, jumping, and steering. 
\textcolor{black}{The pace gait is chosen for its distinct low-speed characteristics, as trials reveal that the easy generalization of the trot gait can hinder the effective learning of jumping behavior at high-speed commands.} 
\textcolor{black}{Next, they are retargeted to the robot's morphology via inverse kinematics, and the corresponding joint pose can be obtained [\ref{peng2020learning}] as dataset $\mathcal{M}$.} To facilitate the algorithm to grasp the difficult jumping skills, the jump data are assigned 6 times higher sampling weights than those of low-speed motions, thereby increasing their probability of being retrieved during training.


\begin{table}[h]
\caption{Reward terms for training low-level policy. \textcolor{black}{(Feet aerial height: feet heights $h^{feet}_{i}$ filtered by the mask ensuring all feet are in the air $m^{air}: \bigcap_{i=1}^{4} \{ F_i < 1.0N \}$).}}
\vspace{-0.8em}
\label{reward_function}
\setlength\tabcolsep{3pt}  
\begin{center}
\begin{tabular}{c c c c}
\hline
\textbf{Term} & \textcolor{black}{\textbf{Ratio}} & \textbf{Equation}  & \textcolor{black}{\textbf{Weight}} \\
\hline
\\ [-1.5ex]
\multirow{3}*{Task $r^{T}_{t}$} & \multirow{3}*{0.8} & $\exp\{-4 (\boldsymbol{v}_{xy}^{cmd}-\boldsymbol{v}_{xy})^2\}$  & 1.0\\ [+0.3ex]
  &   &  $\exp\{-4(\boldsymbol{\omega}^{cmd}_{z}-\boldsymbol{\omega}_{z})^2\}$    & 0.5\\ [+0.3ex]
  &    &  \textcolor{black}{$m^{air} \cdot \sum{_{i=1}^{4}} h^{feet}_{i}$}    & 2.4\\ [+0.3ex]
\hline
\\ [-1.5ex]
Style $r^{S}_{t}$ & 0.2 & $ \max \left[ 0, 1 - 0.25(\mathcal{D}_{\psi} - 1)^2 \right] $ & 1.0\\ [+0.3ex]
\hline
\\ [-1.5ex]
\multirow{3}*{Regularization ${r}^{R}_{t}$} 
     & \multirow{3}*{0.5}
     & $||\boldsymbol{\tau}||^2$  &  $-1 \times 10^{-5}$\\ [+0.3ex]
     & & $||\boldsymbol{a}_{t}-\boldsymbol{a}_{t-1}||^2$ & -0.01\\ [+0.3ex]
     & & $||\boldsymbol{\ddot{q}}||^2$ & $-2.5 \times 10^{-7}$\\ [+0.3ex]
\hline

\end{tabular}
\end{center}
\vspace{-1.5em}
\end{table}

\subsection{Training of High-level Controller} \label{high_level}
\textcolor{black}{Each low-level skill modulates the robot’s whole-body motion differently. Unlike the previous work relying on manual commands to control the locomotion [\ref{wu2023learning}, \ref{vollenweider2023advanced}], we explore the autonomy of this imitation-based policy by incorporating a high-level module. It is responsible for generating the most appropriate commands to traverse the narrow gate (Fig. \ref{pics:framework}). }
The high-level policy is trained in an outer loop running at 10$Hz$, while an inner loop runs the locomotion module at 50$Hz$. The low-level policy of the inner loop is frozen throughout high-level policy training. 

\subsubsection{State space} We reuse locomotion-level states as input to the high-level policy, enabling the agent to access the robot states during high-dynamic motions. 
\textcolor{black}{Therefore, the high-level observation $\boldsymbol{o}^{H}_{t}$ still consists proprioceptive states including $\boldsymbol{\omega}_{t}$, $\boldsymbol{g}_{t}$, $\boldsymbol{q}_{t}$ and $\boldsymbol{\dot{q}}_{t}$ eluded in Section \ref{low_level_state}. }
\textcolor{black}{Apart from these, the action of the last step $\boldsymbol{a}^{H}_{t-1}$ is also included.}
The base linear velocity $\boldsymbol{v}_{t}$ is still part of the privileged observation besides $\boldsymbol{o}^{H}_{t}$.

\begin{figure}[h]
   \centering
   \includegraphics[width=0.51\textwidth, trim=0 0 0 0,clip]{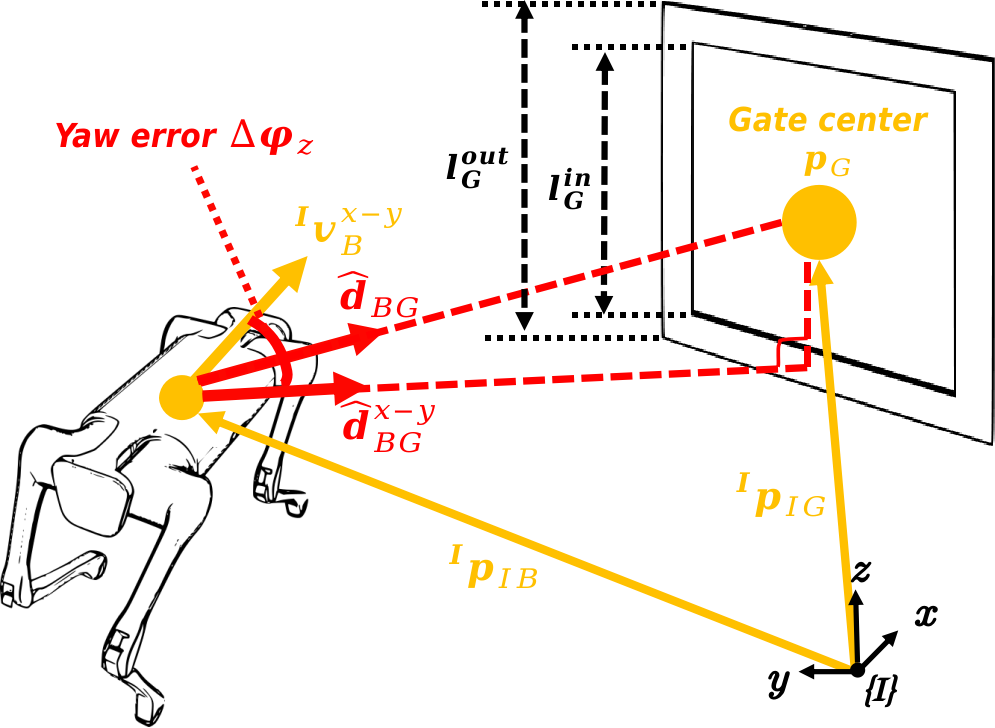}
   \caption{Illustration of the model used in the high-level controller training. The sketch shows the components that determine the relative position of the robot w.r.t the gate. }
   \label{pics:nav_scheme}
\end{figure}

\textcolor{black}{Additionally, the relative position of the gate center w.r.t the robot base is part of the input.} As depicted in Fig. \ref{pics:nav_scheme}, $\widehat{\boldsymbol{d}}_{BG}$ indicating the direction from the robot to the gate center is calculated by:
\begin{equation}
    \widehat{\boldsymbol{d}}_{BG} = 
    \frac{ \prescript{I}{}{\boldsymbol{p}_{IG}} - \prescript{I}{}{\boldsymbol{p}_{IB}}}
    {\Vert \prescript{I}{}{\boldsymbol{p}_{IG}} - \prescript{I}{}{\boldsymbol{p}_{IB}} \Vert}_{2},
\end{equation}
where $\prescript{I}{}{\boldsymbol{p}_{IB}}$ is the pose of the robot base accessed from the external localization technique, and $\prescript{I}{}{\boldsymbol{p}_{IG}}$ is the gap center's pose in the world frame provided by the perceptual detection module (details in Section \ref{perception_module}). Another input of the yaw error $\triangle \boldsymbol{\varphi}_{z}$ indicates the direction deviation:
\begin{equation} \label{eq:delta_yaw}
    \triangle \varphi_{z} = \arccos 
                            \frac{ \widehat{\boldsymbol{d}}^{x-y}_{BG} \cdot \prescript{I}{}{\boldsymbol{v}}^{x-y}_{B} }
                            {\Vert \widehat{\boldsymbol{d}}^{x-y}_{BG} \Vert \cdot \Vert \prescript{I}{}{\boldsymbol{v}}^{x-y}_{B}\Vert},
\end{equation}
where $\widehat{\boldsymbol{d}}^{x-y}_{BG}$ and $\prescript{I}{}{\boldsymbol{v}}^{x-y}_{B}$ are $x-y$ planar projection of $\widehat{\boldsymbol{d}}_{BG}$ and base linear velocity $\prescript{I}{}{\boldsymbol{v}}_{B}$, respectively. Together with the gate geometry $l^{out}_{G}, l^{in}_{G}$ accessible from the perception, $\boldsymbol{o}^{H}_{t}$ is concatenated as:
\begin{equation} \label{eq:nav_input}
\boldsymbol{o}^{H}_{t} = \left[ \boldsymbol{\omega}_{t}, 
                                  \boldsymbol{g}_{t}, 
                                  \boldsymbol{q}_{t},
                                  \boldsymbol{\dot{q}}_{t},
                                  \triangle \varphi_{z},
                                  \widehat{\boldsymbol{d}}_{BG},
                                  \boldsymbol{a}^{H}_{t-1},
                                  l^{out}_{G},
                                  l^{in}_{G},
                            \right]
\end{equation}

\subsubsection{Action space} In essence, the high-level action $\boldsymbol{a}^{H}_{t} \in\mathbb{R}^{2}$ is the linear velocity command in the x-direction $\boldsymbol{v}^{cmd}_{x}$ and the yaw angular velocity command $\boldsymbol{\omega}^{cmd}_{z}$, which are fed to the low-level policy $\pi^{L}$. The locomotion policy then regulates the robot's velocity and orientation by tracking the high-level action command $\boldsymbol{a}^{H}_{t}$ with the goal of optimizing the following reward functions.

\subsubsection{Reward function} The primary objective of this task is to successfully and precisely traverse through the narrow gate's center. Therefore, the robot should move in the direction towards the gate center, i.e. base velocity in the world frame $\prescript{I}{}{\boldsymbol{v}_{B}}$ should align with the vector $\widehat{\boldsymbol{d}}_{BG}$. This is achieved by rewarding the projection of $\prescript{I}{}{\boldsymbol{v}_{B}}$ on the desired direction $\widehat{\boldsymbol{d}}_{BG}$, as well as penalizing $\triangle \varphi_{z}$, the angular difference between $\prescript{I}{}{\boldsymbol{v}}^{x-y}_{B}$ and $\widehat{\boldsymbol{d}}^{x-y}_{BG}$ in Eqn (\ref{eq:delta_yaw}). 

\textcolor{black}{Additionally, the regularization terms encourage less energy dissipation and the exceeding of the command velocity limit.} The detailed reward items of the high-level controller training are listed in Table \ref{nav_reward_function}.

\begin{table}[h]
\caption{Reward terms for training high-level policy}
\vspace{-1.0em}
\label{nav_reward_function}
\setlength\tabcolsep{3pt}  
\begin{center}
\begin{tabular}{c c c}
\hline
\textbf{Term} & \textbf{Equation}  & \textcolor{black}{\textbf{Weight}} \\
\hline
\\ [-2.1ex]
Center tracking &  $\prescript{I}{}{\boldsymbol{v}_{B}} \cdot \widehat{\boldsymbol{d}}_{BG}$    & 2.2\\ [+0.3ex]
Yaw rectification  &  $\triangle \varphi_{z}$ (Eqn. (\ref{eq:delta_yaw}))    & 0.15\\ [+0.3ex]
Energy dissipation & $(\boldsymbol{\tau} \cdot \boldsymbol{\dot{q}})^2$  & -3$\times$$10^{-5}$ \\ [+0.3ex]
Action limits & $\max(0, \boldsymbol{a}^{H}_{t} - \boldsymbol{v}^{max}_{cmd}, \boldsymbol{v}^{min}_{cmd} - \boldsymbol{a}^{H}_{t})$ & -8.0\\[+0.3ex]
\hline

\end{tabular}
\end{center}
\vspace{-2.0em}
\end{table}

\subsection{Perception Module via Onboard Sensing} \label{perception_module}
We adapt our previous gate-detection algorithm [\ref{xie2023learning}] to specifically identify the black square frames in the current experiment, using only an RGB-D camera to capture both color images and depth data.

The algorithm first converts a color image to a binary image and then performs closing operations, Canny edge detection, edge undistortion, and edge grouping consecutively. Next, we apply the Douglas-Peucker and Quickhull algorithm to extract a set of convex hulls with four vertices.
Hitherto, the convex hull $\mathcal{C}$ of a rectangle frame 
can be formulated as a group of four-pixel points:
\begin{equation}
    \mathcal{C} = \left\{ [u_{i}, v_{i}]^{T} | i = 1, ... , 4 \right\}.
\end{equation}
where $[u_{i}, v_{i}]$ is the pixel coordinate of the image. 
Based on this, we can calculate the position of each vertex $\boldsymbol{p}_{v,i} = [x_{v,i},y_{v,i},z_{v,i}]$ in the world frame as:
\begin{equation} \label{eq:camera_transform}
\begin{aligned}
\boldsymbol{\mathcal{P}}_{1} \boldsymbol{\mathcal{P}}_{2} [x_{v,i},y_{v,i},z_{v,i},1]^T = d [u_{i}, v_{i},1]^T,
\end{aligned}
\end{equation}
where $\boldsymbol{\mathcal{P}}_{1} \in \mathbb{R}^{3\times4}$ is the camera intrinsic matrix and $\boldsymbol{\mathcal{P}}_{2} \in \mathbb{R}^{4\times4}$ is the transformation from world to the camera. $d$ is the depth at pixel coordinate $[u_{i}, v_{i}]^T$ and can be obtained by the depth image.
The central position of the gate, denoted as $\boldsymbol{p}_{G}$, is computed as the average of all vertices. Given the gate being a square, the geometrical information of the gate $\boldsymbol{l}_{G} = [l_{G}^{in}, l_{G}^{out}]$ can be calculated as the average length of all edges. To simplify, here we only take the length of the inner square as an example:
\begin{equation}
    l_{G}^{in} = \frac{1}{4}\sum_{i=1}^{4} {\Vert \boldsymbol{p}^{in}_{v,i} - \boldsymbol{p}^{in}_{v,i+1} \Vert }_{2}.
\end{equation}
Furthermore, we use a 3rd-order low-pass filter to smooth the detection output $\boldsymbol{x}_g = [\boldsymbol{p}_G, \boldsymbol{l}_{G}]^T$ and finally input to the controller.

\begin{figure}[h]
   \centering
   \includegraphics[width=0.52\textwidth, trim=0 0 0 0,clip]{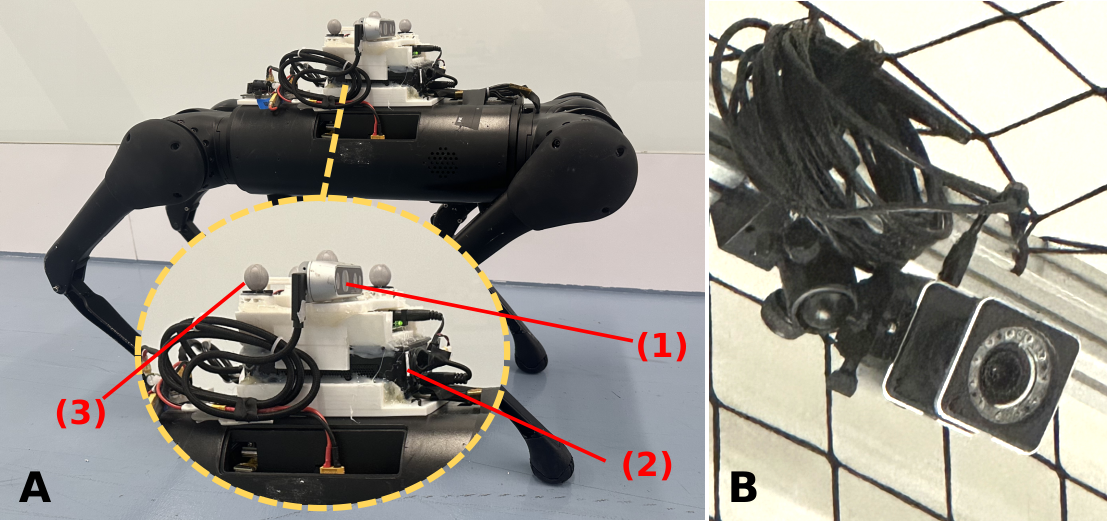}
   \caption{Real-hardware setup: (\textbf{A}) \textcolor{black}{(1) Forward-facing Intel D435i depth camera,} (2) Intel NUC onboard computer, (3) VICON marker, which all mounted on a Unitree Aliengo weighing 22 kg. (\textbf{B}) VICON motion capture (MoCap) system, which provides the robot's pose information. The hardware tests are all conducted in a 6.0m $\times$ 3.0m MoCap venue. }
   \label{pics:hardware_setup}
\end{figure}

\section{Experimental Results and Discussion}\label{Experiments}
\subsection{Implementation details}
\subsubsection{Low-level controller}
We train the low-level policy on 5480 agents in parallel on the Isaac Gym simulator [\ref{makoviychuk2021isaac}] for 25000 episodes. 
The linear velocity command in the forward direction is set as $[0.0, 2.4]m/s$ 
, and the angular velocity command as $[-1.57, 1.57]rad/s$. To enhance policy transferability from simulation to real hardware, we randomize the parameters in our offline-trained actuator net including the motor's strength, latency, and offset (Table \ref{rand_parameters}). 

\textcolor{black}{In the end, for a well-trained policy, motion style transitions are conditioned on varied commanded velocities: low-speed commands primarily trigger a pacing gait ($\boldsymbol{v}^{cmd}_{t} \in [0.25, 1.57]$[m/s]), while high-speed commands are needed to initiate a running jump ($\boldsymbol{v}^{cmd}_{t} \in [1.42, 2.40]$[m/s]). 
This distinctly different motion style is also attributed to the dataset constructed in Section \ref{dataset}, featuring clear speed disparity.
}

\begin{table}[h]
\vspace{-1.0em}
\caption{Randomization range of parameters}
\label{morph_table}
\begin{center}
\vspace{-1.0em}
\begin{tabular}{c c c c}
\hline
\textbf{Controller} & \textbf{Parameters} & \textbf{Range [Min, Max]} &\textbf{Unit}\\
\hline
\\ [-2.1ex]
   \multirow{4}*{Low-level}   & Added mass   & [-2.0, 2.0]   & Kg \\
   &  Motor strength & [0.9, 1.1]   & - \\
   & Motor latency & [0.0, 20.0]  & ms \\
   & Motor offset  & [-0.02, 0.02] & Rad \\

\hline
\\ [-2.1ex]
   \multirow{5}*{High-level}  
   & Gate center $x$ &  [1.2, 6.0] &  m \\
   & Gate center $y$ &   [-1.8, 1.8] &  m \\
   & Gate center $z$ &  [0.5, 0.63] & m \\
   & Gate size $l^{out}_{G}$ &  [0.7, 1.0] &  m \\
   & Gate thickness &  [0.02, 0.1] &  m \\
   \hline

\end{tabular}
\end{center}
\label{rand_parameters}
\vspace{-1.0em}
\end{table}

\subsubsection{High-level controller} The running frequency ratio between the high-level module (10$Hz$) and low-level module (50$Hz$) is guaranteed by a sequential setup in the training process. More specifically, given low-level controller has a constant interval of 0.02$s$, the high-level controller sends an action command $\boldsymbol{a}^{H}_{t}$ only after the low-level controller has taken 5 forward steps and sends motor-level action $\boldsymbol{a}^{L}_{t}$. It is worth mentioning that this sequential scheme is also employed in real-world deployment where we set a constant ROS rate for the low-level forward step.
The gate is randomly initialized at different positions in the world frame to enhance the adaptability with the range shown in Table. \ref{rand_parameters}.

The policy net of the low-level and high-level modules are both 4-layer multi-layer perception (MLP), which has the hidden layer architecture $[512, 256, 128]$ and Exponential Linear Units (ELU) activation. 
The discriminator in the low-level controller training is also an MLP with hidden layers of size $[1024, 512]$ and Leaky Rectified Linear Unit (ReLU) activation layers. The entire training is performed on a desktop PC with an NVIDIA RTX 4080 GPU. Training of low-level policy costs approximately 11.5 hours, and the training of high-level policy converges in 6 hours.

As depicted in Fig. \ref{pics:hardware_setup}, The hardware validation test for the Aliengo quadrupedal robot, which can generate a maximum torque of approximately 45$N m$, is conducted within a motion capture system area. The depth camera and VICON markers provide the perceptual and localization information, respectively. 
The low-level and high-level controllers run synchronously in a single ROS node on the onboard computer. The perception module, namely the gate detection algorithm operates asynchronously in another node, processing the image received from the depth camera. The extracted gate geometry and location information, which compose part of the observation $\boldsymbol{o}^{H}_{t}$, are then passed on to the high-level controller.

\begin{figure*}[ht]
   \centering
    \includegraphics[width=1.0\textwidth, trim=1 3 2 7,clip]{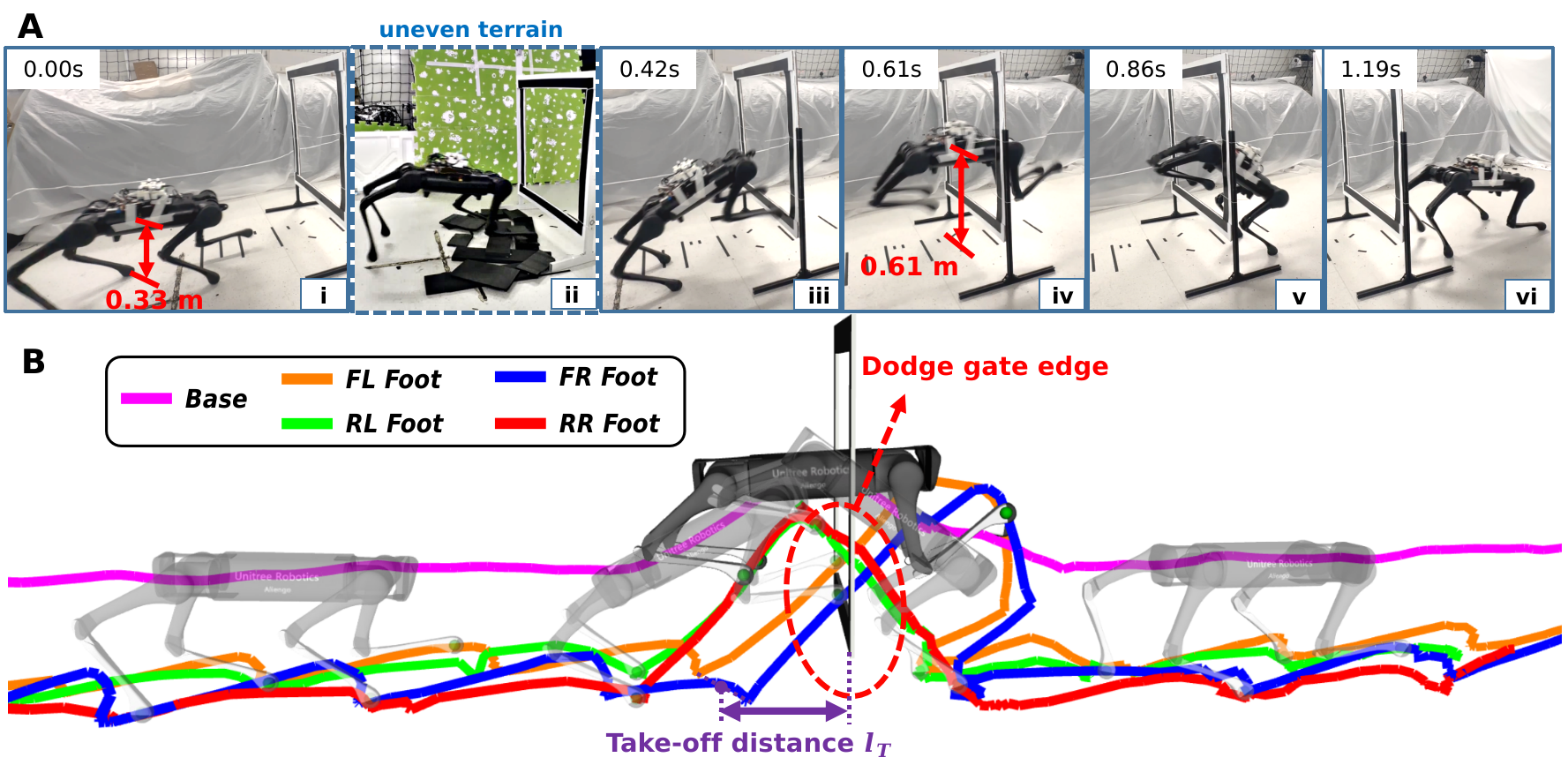}
   \caption{(\textbf{A}) The snapshots are taken from the real deployment of the proposed pipeline on a real Unitree Aliengo robot. The robot detects the gate size and position via onboard sensing. Then, the controllers determine the suitable skills and switching times that enable robots to jump through the gate.
   \textcolor{black}{(ii) showcases the special case when taking off on uneven terrain.} 
   (\textbf{B}) The visualization of the base and foot trajectories in (\textbf{A})'s test. The high-level policy fully leverages the low-level control policy capability, \textcolor{black}{guiding the foot end-effectors to dodge the gate edge autonomously.} }
   \label{pics:vis}
\end{figure*}

\subsection{\textcolor{black}{Autonomous Behavioral Transition}}

\begin{figure}[H]
   \centering
    \includegraphics[width=0.57\textwidth, trim=1 1 1 1,clip]{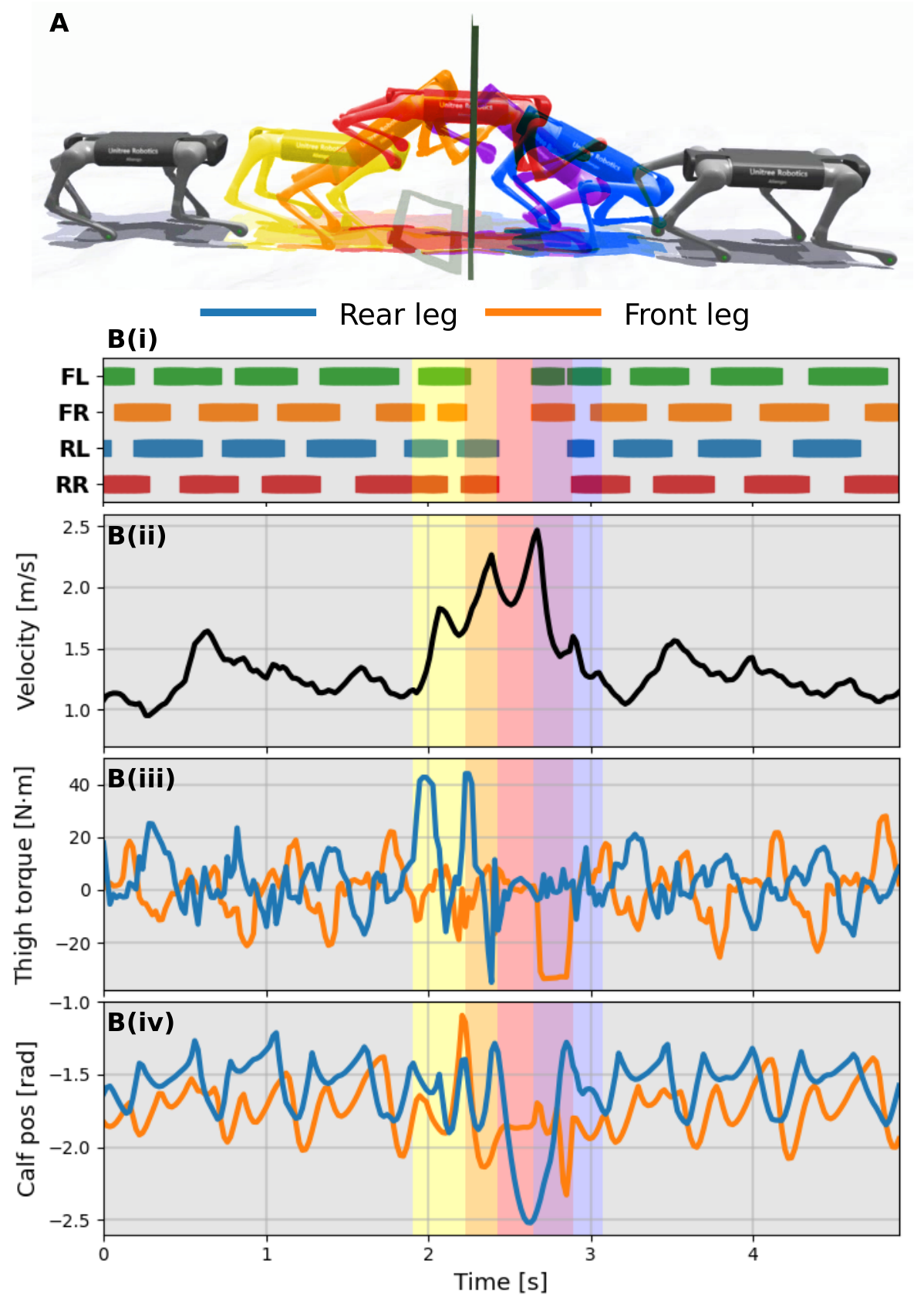}
   \caption{ (\textbf{A}) \textcolor{black}{Illustration of the continuously varied skills during the whole traversal process. The marked colors correspond to the time periods in the below profiles. } (\textbf{B(i)}) Gait. (\textbf{B(ii)}) Base velocity.  (\textbf{B(iii)}) Torque of front and rear thigh motors. (\textbf{B(iv)}) Pose of front and rear calf joints. }
   \label{pics:transition}
   \vspace{-1.0em}
\end{figure}

\textcolor{black}{The system enables the robot to jump through narrow gates by generating suitable maneuvers in a well-coordinated manner, as depicted in Fig. \ref{pics:transition}.
}
\textcolor{black}{When the robot is at a distance from the gate, it uses a slow-pacing gait to conserve energy.}
\textcolor{black}{As it approaches the gate (\textbf{Yellow} in Fig. \ref{pics:transition} (A)), a large stride requires a surge of torque at the rear leg's motors.}
\textcolor{black}{The motors are nearly saturated when the robot takes off (\textbf{Orange}).
After the robot is accelerated to a maximum speed of up to 2.5 $m/s$,
fast deacceleration is observed (\textbf{Purple}) leading to the high torque at the front motor to catch the fall of the robot.}
\textcolor{black}{It is worthwhile to note that there is no pre-defined motion or contact sequence like the traditional methods.} 
\textcolor{black}{The system relies solely on the scenario at hand, to determine which maneuvers to execute.}

\vspace{-0.5em}

\textcolor{black}{Furthermore, the system coordinates all joints in real time to generate feasible whole-body trajectories, as visualized in Fig. \ref{pics:vis}. The planned trajectory of the feet's end-effectors autonomously dodges the gate edge when the rear calf joints deflect by more than 140$^{\circ}$. This indicates that the control system is capable of exploiting a large portion of the motor range to circumvent collision. The ``airborne time" in this process is close to 0.44$s$.}

\subsection{\textcolor{black}{Comparative Studies}} \label{section: ablation}
\begin{figure}[H]
    \centering
    \includegraphics[width=0.7\textwidth, trim=0 0 0 0,clip]{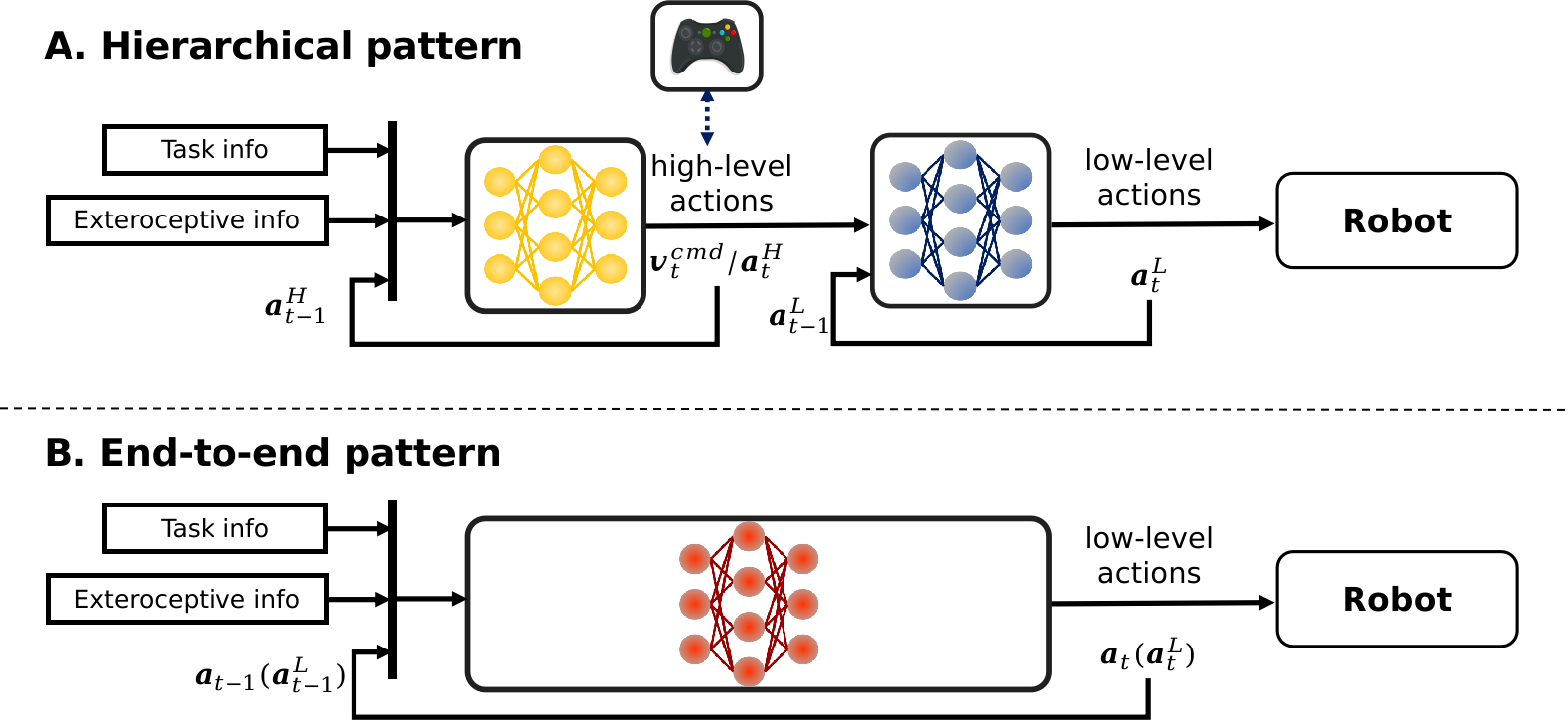}
    \caption{\textcolor{black}{Hierarchical (A) and End-to-end (B) control architecture used for implementations.} }  
    \label{fig: compare_arch}
\end{figure}

To evaluate the advantage of the hierarchical architecture, \textcolor{black}{we consider the following ablated training settings and existing state-of-the-art algorithms}:

\begin{itemize}
    \item \textbf{E2EStyle}: 
    \textcolor{black}{The baseline shares identical exteroceptive inputs and proprioceptive states with our method. The only difference is the absence of hierarchical signals (velocity commands  $\boldsymbol{v}_{t}^{cmd}$, and high-level actions $\boldsymbol{a}^{H}_{t-1}$), as these are inherently incompatible with end-to-end architectures (Fig. \ref{fig: compare_arch}). The policy outputs directly the joint-level actions.}
    All task relevant rewards of Table \ref{reward_function} and \ref{nav_reward_function} are leveraged.
    \item \textbf{E2EBase}: 
    \textcolor{black}{The observation is aligned with as above.} This end-to-end policy variant ablates the style reward, which is solely guided by task-specific rewards and completely disregards imitation. \textcolor{black}{We refer the reader to the details of reward terms demonstrated in Table \ref{tab:baseline_setting_rew2}.}
    \textcolor{black}{\item \textbf{Parkour} [\ref{cheng2023extreme}]: Another E2E policy that conquers obstacles agilely using the scandots of the undulating terrain and a specified reward for moving toward the target.}
    \item \textcolor{black}{\textbf{ConsJump} (Ours): The proposed hierarchical architecture with the low-level skills as continuous states.}
    \textcolor{black}{\item \textbf{HierGate} [\ref{yang2020multi}]: This pipeline differs from ours by training discrete low-level skills and a high-level gate network to assign their respective weights.}
    \textcolor{black}{\item \textbf{HierDisc} [\ref{hoeller2024anymal}]: This pipeline differs from ours by training discrete low-level skills and a one-hot encoding style high-level policy to select them.}

\end{itemize}

\begin{table}[h]
\caption{\textcolor{black}{Reward settings for the training of different methods}}
\vspace{-0.5em}
\label{tab:baseline_setting_rew2}
\setlength\tabcolsep{3pt}  
\begin{center}
\begin{tabular}{c c c c c}
\hline
\textbf{\textcolor{black}{Reward Term}} & \textbf{\textcolor{black}{E2E-BASE}}  & \textbf{\textcolor{black}{E2E-STYLE}}& \textbf{\textcolor{black}{Hier High-level}}& \textbf{\textcolor{black}{Hier Low-level}}\\
\hline
\\ [-2.1ex]
\textcolor{black}{Center tracking}           &  \textcolor{black}{$\surd$}    &  \textcolor{black}{$\surd$}   & \textcolor{black}{$\surd$}   & \textcolor{black}{$\times$} \\ [+0.3ex]
\textcolor{black}{Yaw rectification}         &  \textcolor{black}{$\surd$}    &  \textcolor{black}{$\surd$}   & \textcolor{black}{$\surd$}   & \textcolor{black}{$\times$} \\ [+0.3ex]
\textcolor{black}{Energy dissipation}        &  \textcolor{black}{$\surd$}    &  \textcolor{black}{$\surd$}   & \textcolor{black}{$\surd$}   & \textcolor{black}{$\times$} \\ [+0.3ex]
\textcolor{black}{Command action limits}             &  \textcolor{black}{$\times$}   &  \textcolor{black}{$\times$}  & \textcolor{black}{$\surd$}   & \textcolor{black}{$\times$} \\ [+0.3ex]
\textcolor{black}{Velocity tracking}         &  \textcolor{black}{$\times$}   &  \textcolor{black}{$\times$}  & \textcolor{black}{$\times$}  & \textcolor{black}{$\surd$}  \\ [+0.3ex]
\textcolor{black}{Style reward}              &  \textcolor{black}{$\times$}   &  \textcolor{black}{$\surd$}   & \textcolor{black}{$\times$}  & \textcolor{black}{$\surd$}  \\ [+0.3ex]
\textcolor{black}{Regularization}            &  \textcolor{black}{$\surd$}    &  \textcolor{black}{$\surd$}   & \textcolor{black}{$\times$}  & \textcolor{black}{$\surd$}  \\ [+0.3ex]
\hline
\end{tabular}
\end{center}
\end{table}

\begin{figure}[H]
   \centering
\includegraphics[width=0.79\textwidth]{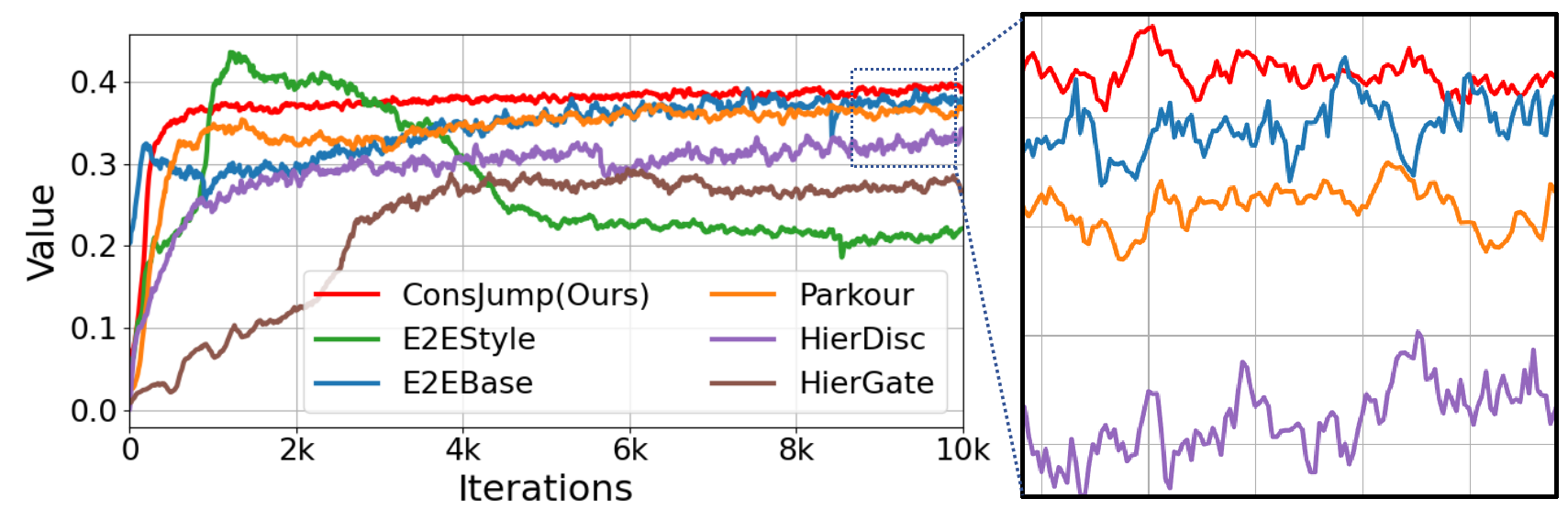}
   \caption{ Learning curves of compared algorithms. 
   Because of differences in reward setups between [\ref{cheng2023extreme}] and ours, we recalculate the task rewards during training and log them for comparison only. 
    }
   \label{pics:baseline}
\end{figure}

\begin{figure}[H]
   \centering
\includegraphics[width=0.73\textwidth, trim=1 1 1 6,clip]{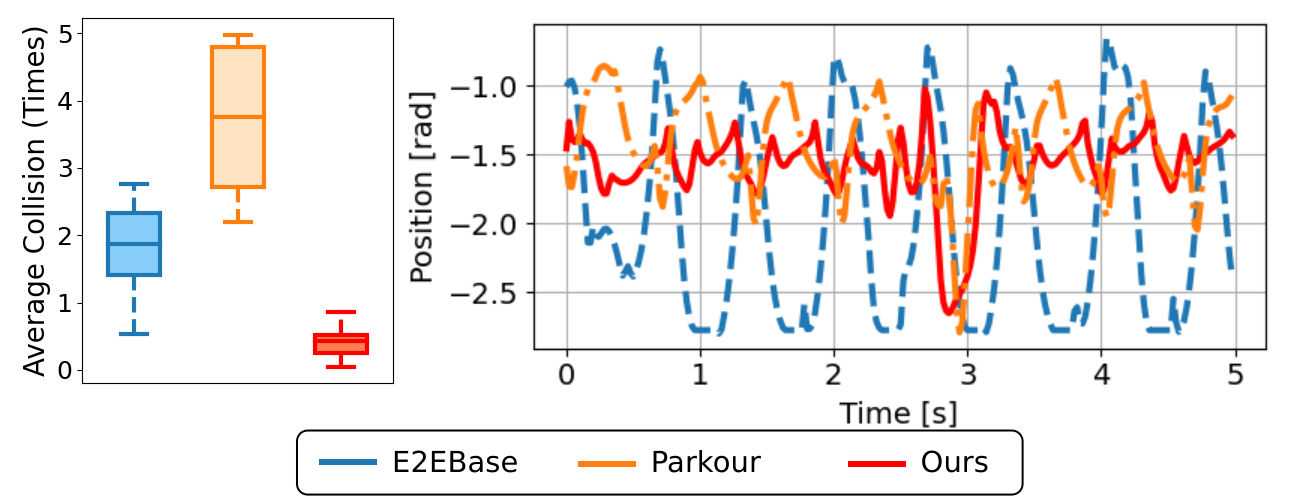}
   \caption{ The average collision (Left) and measured positions of the calf joint (Right) during the traversal process.    }
   \label{pics:joint_baseline}
   \vspace{-0.6em}
\end{figure}

\textcolor{black}{As depicted in Fig. \ref{pics:baseline}, among three controllers that achieve high task rewards, the highly task-driven E2EBase policy outputs a jerky joint position command, causing the robot to jump aggressively (Fig. \ref{pics:joint_baseline}).} 
\textcolor{black}{Notably, Fig. \ref{pics:joint_baseline} illustrates that baseline methods lead to significantly more collisions during locomotion, along with more violent joint angle oscillations compared to our approach.}
The sparsity of heightmap features results in the parkour policy, suitable for traversing large bulk of terrain, operate in a 'tripping over' manner. \footnote{\textcolor{black}{We also test an alternative baseline variant where the Parkour observation is augmented with explicit geometric information about the constrained obstacle. Its performance remains comparable to the vanilla Parkour implementation.} }
E2EStyle performance plummets with more iterations as the combination of imitation and this complex task makes the training highly unstable. 
\textcolor{black}{The fundamental limitation is that these learn-from-scratch approaches inherently require extensive exploration across a vast action space to discover viable solutions for specific tasks. This necessitates meticulous reward shaping to guide the policy toward desired behaviors, such as jumping, while avoiding unsafe or inefficient motions.}

\textcolor{black}{In contrast, our hierarchical framework leverages a pre-trained low-level policy with a fixed repertoire of dynamic skills rather than learning motions from scratch. The high-level policy only needs to master the selection and sequencing of these pre-optimized skills to accomplish the task. 
This structure offers key advantages that the high-level policy explores a smaller, semantically meaningful action space instead of raw joint-level actions, significantly reducing training complexity.}
Our pipeline consistently outperforms all the other methods including the HierDisc where the non-smooth selection leads to a lack of coherence during controller switching, making it difficult to pinpoint the optimal take-off point.
We refer the reader to the linked video for more comparison details.

\textcolor{black}{While our hierarchical framework simplifies the high-level module by abstracting low-level skills through velocity commands, it may limit the granularity of joint-level control compared to end-to-end methods like Parkour \ref{cheng2023extreme}. Specifically, distinct joint configurations with similar velocity profiles (e.g., different leg retraction strategies during jumping) cannot be explicitly selected by the high-level policy. However, in our task, the imitation-based low-level policy inherently prioritizes biomechanically efficient poses (e.g., tucked legs during aerial phases), aligning with the optimal joint configurations observed in animal behaviors. This design choice trades off fine-grained joint control for robustness and training efficiency, which is critical for real-world deployment.}

\vspace{-0.5em}

\subsection{Evaluation of System Adaptability} \label{section: adapt}
\begin{figure}[H]
   \centering
\includegraphics[width=1.0\textwidth, trim=1 1 4 1,clip]{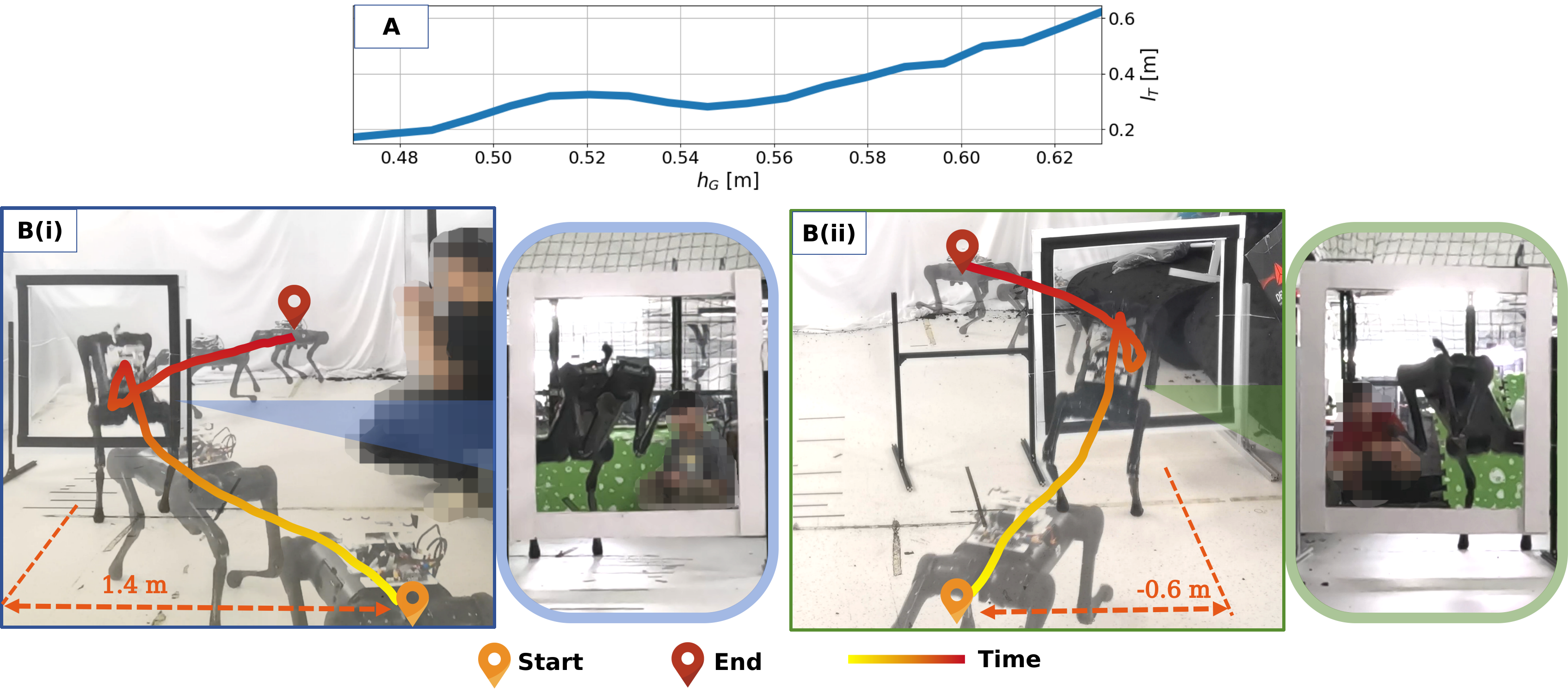}
   \caption{\textcolor{black}{ConsJump is adaptive to scenarios with varying gate positions. (\textbf{A}) The function of takeoff distances $l_{T}$ in relation to gate heights $h_{G}$.} (\textbf{B}) The full base trajectories and the front views are demonstrated. \textcolor{black}{Please refer to the linked video for more tests on gate position adaptability.} }
   \label{pics:real_nav}
   \vspace{-0.8em}
\end{figure}

\begin{figure}
   \centering
\includegraphics[width=1.0\textwidth]{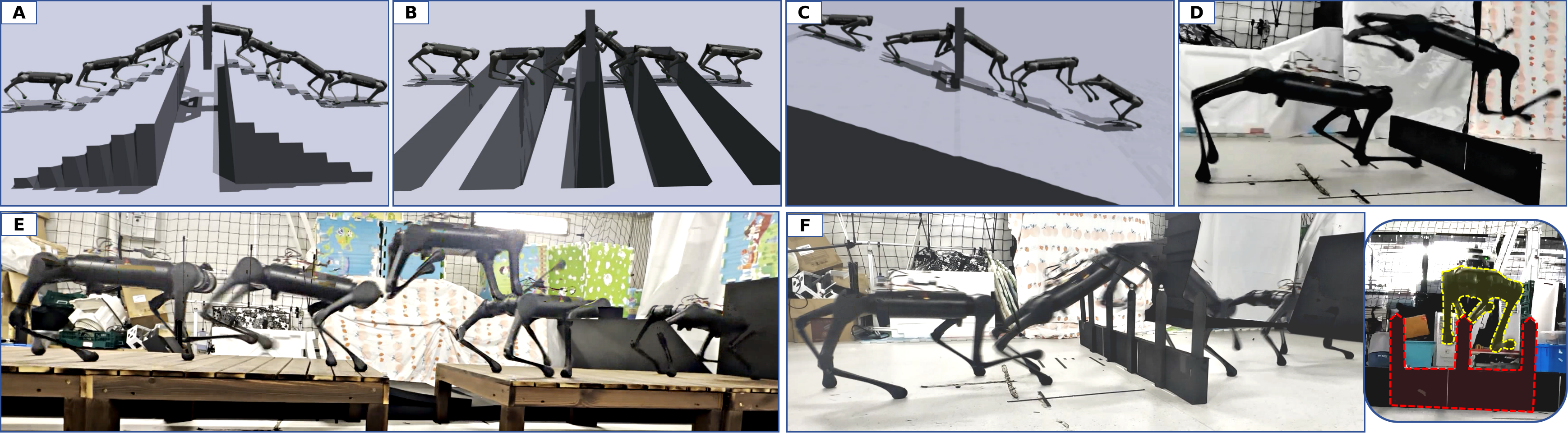}
   \caption{ The system's generalization to various complex terrains: traversing the gate on continuous stairs (\textbf{A}), continuous gaps (\textbf{B}) and slope (\textbf{C}). The system's extension to various tasks: jump over a hurdle (\textbf{C}), gap (\textbf{E}), and fence with boards arranged vertically in a row (\textbf{F}, with the front view). The accurate shape information in the real-world tests is accessed via the VICON system.  }
   \label{pics:terr}
\end{figure}

\textcolor{black}{It is demonstrated in Fig. \ref{pics:real_nav} and Fig. \ref{pics:terr} (A $\sim$ C) that the system further has the ability to adapt to changeable environments, particularly the complex terrains, and varied gate positions. } 
\textcolor{black}{The task requires jumping over the gate while conquering complex terrains, such as climbing up and down stairs and stepping over gaps.}
\textcolor{black}{Even in cases of foot sliding on uneven terrain before takeoff, the robot can still accurately accomplish the traverse, as depicted in Fig. \ref{pics:vis} (A)(ii) and the video. }

Compared to the scenario of reduced gate height, when traversing a taller one, \textcolor{black}{the controller needs to allocate a longer take-off distance in advance and make more aggressive adjustments to the joints' motion (Fig. \ref{pics:real_nav} (A))}. 
\textcolor{black}{The robot can also be steered to jump diagonally over a deviated gate, which has a smaller facing area.} As the motion capture venue has a limited lateral length, the experiment starts with the gate positioned at 1.4 m in the y direction (Fig. \ref{pics:real_nav} (B)) and is repeated twice every 0.4 m until a position of -1.4 m is reached, i.e. 
$y \in \{ 1.4 - 0.4 k \ | \ k=0, .. , 7 \}$.
In all trials, the resulting controller accomplishes the task without a single failure. 

\begin{figure}[t]
   \centering
    \includegraphics[width=1.0\textwidth, trim=1 1 1 1,clip]{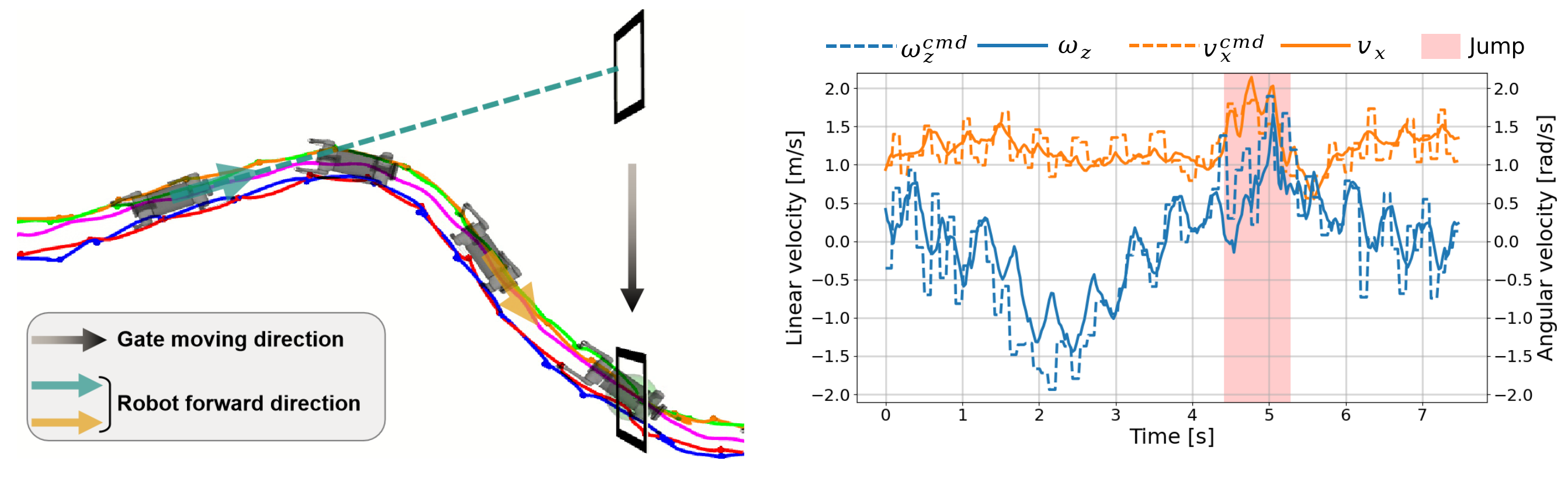}
   \caption{When the gate moves dynamically, the robot can identify the change and adapt its trajectory accordingly. The velocity command given by the high-level controller during the experiment is presented below. }
   \label{pics:dynamic_tracking}
\end{figure}

We delve deeper into the adaptability of emergent changes. The system exhibits reactive behaviors against the sudden shift in the gate location, as illustrated in Fig. \ref{pics:dynamic_tracking}.
The velocity command outputted by the high-level policy changes correspondingly. This guides the base in maintaining its orientation towards the gate center and ultimately completing the traversal. \textcolor{black}{The low-level controller shows nice overall tracking of the dynamically varied commands with the average linear and angular velocity tracking errors being 0.0258 m/s and 0.0857 rad/s, respectively}.

Lastly, as shown in Fig. \ref{pics:terr} (D $\sim$ F), it is verified that the proposed hierarchical architecture is advantageous because it can be easily extended to other specific tasks by reconfiguration of high-level modules, saving the effort of retraining low-level policy. 
In the highly constrained scenario (Fig. \ref{pics:terr} (F)), the reconfigured high-level module, by incorporating the positions of adjacent vertical obstacles and the lower threshold height, guides the quadruped's leg to pass exactly through the space between them.

\section{CONCLUSIONS}\label{conclutions and future works}

Our work demonstrates a learning-based legged robot system for highly dynamic jumping through confined spaces. 
In this hierarchical framework,
the low-level locomotion is obtained via imitation learning, enabling the tracking controller to emulate animal-like dynamic behaviors, and the high-level controller, receiving the detection information from a perception module, determines the appropriate motion sequence for task completion. 
We validate the adaptability of the system and the superiority of the hierarchy via a series of tests in simulation and real hardware.
The research demonstrates that the legged robots have the potential to match the agility of legged animals in high-dynamic scenarios.

\textcolor{black}{While the velocity-based skill abstraction enables robust traversal of narrow obstacles, it inherently constrains the policy from exploiting task-specific joint coordination patterns beyond those embedded in the imitation dataset. This limitation could become prominent when tackling scenarios requiring non-biological motion profiles. Future work could explore hybrid architectures that combine hierarchical skill selection with targeted joint-level adjustments.} A natural extension of this system incorporates a more advanced visual perception, which enables the robot to identify and localize the occlusion more efficiently.
Another promising exploration is incorporating more maneuvers into the repertoire for conquering other constrained workspaces.

\medskip
\textbf{Supporting Information} \par 
Supporting Information is available from the Wiley Online Library or from the author.


\medskip

%

\textbf{References}\\

\refstepcounter{refcounter}\label{nguyen2019optimized} 1 ((Conference articles)) Q. Nguyen, M. J. Powell, B. Katz, J. Di Carlo, S. Kim, presented at Int. Conf. Robot. Autom. (ICRA), 2019.

\refstepcounter{refcounter}\label{Park_2015} 2 ((Conference articles)) H.-W. Park, P. Wensing, S. Kim, presented at Robot.: Sci. Syst., Rome, Italy, July, 2015.

\refstepcounter{refcounter}\label{ding2024robust} 3 ((Journal articles)) J. Ding, V. Atanassov, E. Panichi, J. Kober, C. Della Santina, IEEE Trans. Robot. 2024.

\refstepcounter{refcounter}\label{li2024cafempc} 4 ((Journal articles)) H. Li, P. M. Wensing, IEEE Trans. Robot. 2024.

\refstepcounter{refcounter}\label{roscia2023orientation} 5 ((Journal articles)) F. Roscia, A. Cumerlotti, A. Del Prete, C. Semini, M. Focchi, Sensors, vol. 23, no. 3, p. 1234, 2023.

\refstepcounter{refcounter}\label{cheng2023extreme} 6 ((Conference articles)) X. Cheng, K. Shi, A. Agarwal, D. Pathak, presented at Int. Conf. Robot. Autom. (ICRA), 2024.

\refstepcounter{refcounter}\label{margolis2022learning} 7 ((Conference articles)) G. B. Margolis, T. Chen, K. Paigwar, X. Fu, D. Kim, S. Kim, P. Agrawal, presented at Conf. Robot Learn. (CoRL), 2022.

\refstepcounter{refcounter}\label{bellegarda2020robust} 8 ((Preprints)) G. Bellegarda, C. Nguyen, Q. Nguyen, arXiv preprint arXiv:2011.07089, 2020.

\refstepcounter{refcounter}\label{yang2023cajun} 9 ((Conference articles)) Y. Yang, G. Shi, X. Meng, W. Yu, T. Zhang, J. Tan, B. Boots, presented at Conf. Robot Learn. (CoRL), 2023.

\refstepcounter{refcounter}\label{atanassov2024curriculum} 10 ((Journal articles)) V. Atanassov, J. Ding, J. Kober, I. Havoutis, C. Della Santina, IEEE Robot. Autom. Magazine, 2024.

\refstepcounter{refcounter}\label{grandia2023doc} 11 ((Journal articles)) R. Grandia, F. Farshidian, E. Knoop, C. Schumacher, M. Hutter, M. B¨acher, ACM Trans. Graph., vol. 42, no. 4, pp. 1–14, 2023.

\refstepcounter{refcounter}\label{zhang2018mode} 12 ((Journal articles)) H. Zhang, S. Starke, T. Komura, J. Saito, ACM Trans. Graph., vol. 37, no. 4, pp. 1–11, 2018.

\refstepcounter{refcounter}\label{zhang2024learning} 13 ((Preprints)) C. Zhang, J. Sheng, T. Li, H. Zhang, C. Zhou, Q. Zhu, R. Zhao, Y. Zhang, L. Han, arXiv preprint arXiv:2402.13473, 2024.

\refstepcounter{refcounter}\label{smith2023learning} 14 ((Conference articles)) L. M. Smith, J. C. Kew, T. Li, L. Luu, X. B. Peng, S. Ha, J. Tan, S. Levine, presented at Robot.: Sci. Syst., 2023.

\refstepcounter{refcounter}\label{kareer2023vinl} 15 ((Conference articles)) S. Kareer, N. Yokoyama, D. Batra, S. Ha, J. Truong, presented at IEEE Int. Conf. Robot. Autom. (ICRA), 2023.

\refstepcounter{refcounter}\label{bohez2022imitate} 16 ((Preprints)) S. Bohez, S. Tunyasuvunakool, P. Brakel, F. Sadeghi, L. Hasenclever, Y. Tassa, E. Parisotto, J. Humplik, T. Haarnoja, R. Hafner, et al., arXiv preprint arXiv:2203.17138, 2022.

\refstepcounter{refcounter}\label{yang2020multi} 17 ((Journal articles)) C. Yang, K. Yuan, Q. Zhu, W. Yu, Z. Li, Sci. Robot., vol. 5, no. 49, p. eabb2174, 2020.

\refstepcounter{refcounter}\label{hoeller2024anymal} 18 ((Journal articles)) D. Hoeller, N. Rudin, D. Sako, M. Hutter, Sci. Robot., vol. 9, no. 88, p. eadi7566, 2024.

\refstepcounter{refcounter}\label{caluwaerts2023barkour} 19 ((Preprints)) K. Caluwaerts, A. Iscen, J. C. Kew, W. Yu, T. Zhang, D. Freeman, K.-H. Lee, L. Lee, S. Saliceti, V. Zhuang, et al., arXiv preprint arXiv:2305.14654, 2023.

\refstepcounter{refcounter}\label{peng2022ase} 20 ((Journal articles)) X. B. Peng, Y. Guo, L. Halper, S. Levine, S. Fidler, ACM Trans. Graph., vol. 41, no. 4, pp. 1–17, 2022.

\refstepcounter{refcounter}\label{mellinger2012trajectory} 21 ((Journal articles)) D. Mellinger, N. Michael, V. Kumar, Int. J. Robot. Res., vol. 31, no. 5, pp. 664–674, 2012.

\refstepcounter{refcounter}\label{kaufmann2023champion} 22 ((Journal articles)) E. Kaufmann, L. Bauersfeld, A. Loquercio, M. M¨uller, V. Koltun, D. Scaramuzza, Nature, vol. 620, no. 7976, pp. 982–987, 2023.

\refstepcounter{refcounter}\label{zhuang2023robot} 23 ((Conference articles)) Z. Zhuang, Z. Fu, J. Wang, C. G. Atkeson, S. Schwertfeger, C. Finn, H. Zhao, presented at Conf. Robot Learn. (CoRL), 2023.

\refstepcounter{refcounter}\label{li2023autonomous} 24 ((Journal articles)) Z. Li, J. Zeng, S. Chen, K. Sreenath, Int. J. Robot. Res., vol. 42, no. 8, pp. 565–585, 2023.

\refstepcounter{refcounter}\label{kim2024not} 25 ((Journal articles)) Y. Kim, H. Oh, J. Lee, J. Choi, G. Ji, M. Jung, D. Youm, J. Hwangbo, IEEE Trans. Robot., 2024.

\refstepcounter{refcounter}\label{gilroy2021autonomous} 26 ((Conference articles)) S. Gilroy, D. Lau, L. Yang, E. Izaguirre, K. Biermayer, A. Xiao, M. Sun, A. Agrawal, J. Zeng, Z. Li, et al., presented at IEEE Int. Conf. Autom. Sci. Eng. (CASE), 2021.

\refstepcounter{refcounter}\label{luo2024moral} 27 ((Journal articles)) Z. Luo, Y. Dong, X. Li, R. Huang, Z. Shu, E. Xiao, P. Lu, IEEE Robot. Autom. Lett., vol. 9, no. 5, pp. 4019–4026, 2024.

\refstepcounter{refcounter}\label{peng2021amp} 28 ((Journal articles)) X. B. Peng, Z. Ma, P. Abbeel, S. Levine, A. Kanazawa, ACM Trans. Graph., vol. 40, no. 4, pp. 1–20, 2021.

\refstepcounter{refcounter}\label{escontrela2022adversarial} 29 ((Conference articles)) A. Escontrela, X. B. Peng, W. Yu, T. Zhang, A. Iscen, K. Goldberg, P. Abbeel, presented at IEEE/RSJ Int. Conf. Intell. Robots Syst. (IROS), 2022.

\refstepcounter{refcounter}\label{peng2020learning} 30 ((Conference articles)) X. B. Peng, E. Coumans, T. Zhang, T.-W. Lee, J. Tan, S. Levine, presented at Robot.: Sci. Syst., 2020.

\refstepcounter{refcounter}\label{wu2023learning} 31 ((Journal articles)) J. Wu, G. Xin, C. Qi, Y. Xue, IEEE Robot. Autom. Lett., 2023.

\refstepcounter{refcounter}\label{vollenweider2023advanced} 32 ((Conference articles)) E. Vollenweider, M. Bjelonic, V. Klemm, N. Rudin, J. Lee, M. Hutter, presented at IEEE Int. Conf. Robot. Autom. (ICRA), 2023.

\refstepcounter{refcounter}\label{xie2023learning} 33 ((Journal articles)) Y. Xie, M. Lu, R. Peng, P. Lu, IEEE Robot. Autom. Lett., vol. PP, pp. 1–8, 2023.

\refstepcounter{refcounter}\label{makoviychuk2021isaac} 34 ((Preprints)) V. Makoviychuk, L. Wawrzyniak, Y. Guo, M. Lu, K. Storey, M. Macklin, D. Hoeller, N. Rudin, A. Allshire, A. Handa, et al., arXiv:2108.10470, 2021.

\end{document}